%% file: main.tex
\documentclass[10pt,twocolumn,letterpaper]{article}

\usepackage{cvpr}              


\usepackage{multirow}
\usepackage{float}

\definecolor{cvprblue}{rgb}{0.21,0.49,0.74}
\usepackage[pagebackref,breaklinks,colorlinks,allcolors=cvprblue]{hyperref}
\usepackage{gensymb}
\def\paperID{*****} 
\def\confName{CVPR}
\def\confYear{2025}

\title{SemNav: A Model-Based Planner for Zero-Shot Object Goal Navigation Using Vision-Foundation Models}

\author{{Arnab Debnath}, {Gregory J. Stein}, {Jana Ko{\v{s}}eck{\'a}}\\
George Mason University\\
Fairfax, VA\\
{\tt\small adebnath@gmu.edu, gjstein@gmu.edu, kosecka@gmu.edu}
}

\begin{document}
\maketitle

\input{sec/0_abstract}    
\input{sec/1_intro}

\input{sec/2_relatedWork}
\input{sec/4_approach}
\input{sec/5_experiments}
\input{sec/6_result}

\input{sec/7_conclusion}

\input{}
{
    \small
    \bibliographystyle{ieeenat_fullname}
    \bibliography{main}
}
\input{sec/X_suppl}

\end{document}

%% file: sec/0_abstract.tex
\begin{abstract}
Object goal navigation is a fundamental task in embodied AI, where an agent is instructed to locate a target object in an unexplored environment. Traditional learning-based methods rely heavily on large-scale annotated data or require extensive interaction with the environment in a reinforcement learning setting, often failing to generalize to novel environments and limiting scalability. To overcome these challenges, we explore a zero-shot setting where the agent operates without task-specific training, enabling more scalable and adaptable solution. Recent advances in Vision Foundation Models (VFMs) offer powerful capabilities for visual understanding and reasoning, making them ideal for agents to comprehend scenes, identify relevant regions, and infer the likely locations of objects. In this work, we present a zero-shot object goal navigation framework that integrates the perceptual strength of VFMs with a model-based planner that is capable of long-horizon decision making through frontier exploration. We evaluate our approach on the HM3D dataset using the Habitat simulator and demonstrate that our method achieves state-of-the-art performance in terms of \textit{success weighted by path length} for zero-shot object goal navigation.
\end{abstract}

%% file: sec/1_intro.tex
\vspace{-1.5em}
\section{Introduction}
\label{sec:intro}
Object-goal navigation is a core challenge in embodied AI, where an agent is tasked with locating and navigating to an instance of a specified object within an unfamiliar environment. Unlike map-based or point-goal navigation, this task requires both semantic understanding and active perception, as the agent must interpret visual cues to identify the target object without prior knowledge of its location. This capability is essential for a wide range of real-world applications, including service robots in domestic environments, assistive robots for the elderly, and search-and-rescue operations in disaster-stricken areas. 

Over the years, various learning-based methods have been proposed for object-goal navigation. Supervised approaches \cite{ramakrishnan2022poni} rely on large-scale annotated datasets, while reinforcement learning (RL) techniques \cite{chaplot2020object, pal2021learning, yadav2023ovrl} learn policies through trial-and-error guided by reward shaping. However, these methods are limited to closed sets of object categories and are tightly coupled to their training environments. Extending them to new objects or environments often requires costly retraining, hindering scalability and real-world deployment where generalization to unseen categories and dynamic layouts is essential.


Recent advances in VFMs have greatly enhanced the generalization capabilities of perception systems. Open-vocabulary detectors \cite{li2022grounded, liu2024grounding} and segmentation models \cite{dong2023maskclip, kirillov2023segment} can recognize diverse object categories without a fixed training set, while vision-language models (VLM) and large language models (LLM) provide strong commonsense reasoning about objects and their spatial contexts. VLFM \cite{yokoyama2023vlfm} leverages BLIP-2 \cite{li2023blip2} to compute the cosine similarity between the current observation and a target image, guiding the agent toward visually similar regions. ESC \cite{zhou2023esc} utilizes GLIP \cite{li2022grounded} for detecting candidate objects and rooms, and employs an LLM to reason about the likelihood of the target object being present in a specific room or near certain objects. While these approaches show promise, their reliance on image-text matching or reasoning explicitly with discovered objects can hinder exploration when the target object is far away, occluded, or when the detection is poor. 
In this work, we explore the advancements in VFMs that enables us to use more general prompts which can encourage the agent to move through a hallway or enter a nearby room when the object is not currently visible.

\begin{figure*}[t]
    \centering
    \includegraphics[width=\linewidth]{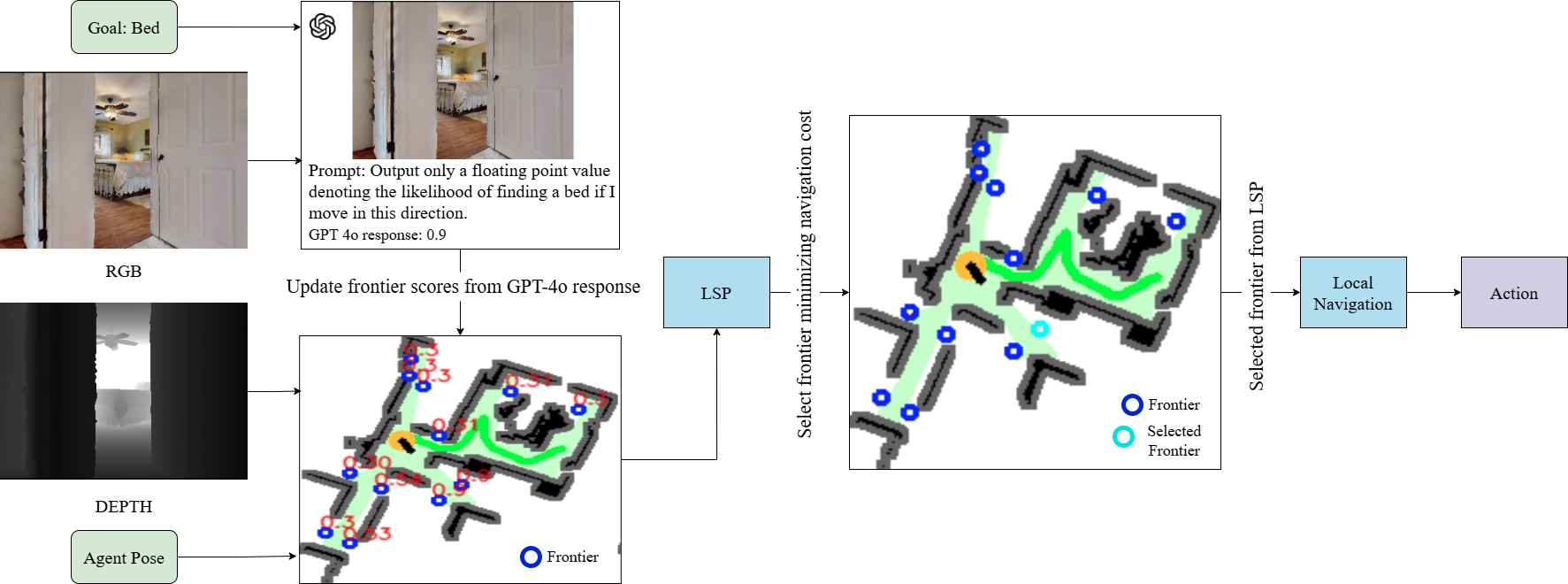}
    \caption{Overview of SemNav }
    \label{fig:overview}
\vspace{-1em}
    
\end{figure*}

Our approach adopts a frontier-based exploration strategy, treating frontiers as long-term subgoals. We leverage VLM to assign the likelihood of finding the target object to these frontiers, guiding the agent’s search. 
Many existing frontier-based methods are inherently myopic: they score each frontier using a heuristic and greedily select the one with the highest score, often leading to suboptimal and inefficient behavior in complex environments. To address this, we build upon the  model-based frontier planning strategy, Learning Over Subgoals Planner (LSP) \cite{stein2018learning}, which evaluates the expected navigation cost of selecting a frontier given the current belief about the environment. This allows the agent to reason about the long-term consequences of its actions and supports more efficient long-horizon planning. 

Our contributions are summarized below:
\begin{itemize}
    \item We propose a semantic frontier-based exploration framework, SemNav, that leverages VLM to assign cost to frontiers for zero-shot object-goal navigation in unseen environments.
    \item We integrate a non-greedy, model-based frontier planning strategy (LSP) to reason about long-term exploration costs, enabling more efficient and informed decision-making compared to traditional greedy frontier selection methods.
\end{itemize}

%% file: sec/2_relatedWork.tex
\vspace{-0.5em}
\section{Related Work}
\vspace{-0.5em}

Zero-shot object-goal navigation leverages VLM and LLMs to enable open-vocabulary object understanding and semantic reasoning without task-specific training. Recent works explore various strategies, including transfer learning \cite{al2022zero, majumdar2022zson}, frontier scoring with VLM and LLMs \cite{yokoyama2023vlfm, yu2023l3mvn}, structured scene reasoning \cite{zhou2023esc}, and topological exploration guided by LLM \cite{wu2024voronav}.

%% file: sec/4_approach.tex
\section{Approach}
An agent is spawned at a random location within an unknown environment and receives a target object category (e.g., couch). 
At each time step, the agent receives visual observations as egocentric RGB-D images and utilizes a GPS+Compass sensor.
The agent can take several actions at each step: \texttt{MOVE\_FORWARD, TURN\_LEFT, TURN\_RIGHT, or STOP}. The \texttt{MOVE\_FORWARD} action moves the agent forward by ${0.25 m}$, and
the \texttt{TURN} actions rotate the agent by $30\degree$ in the specified direction.  An episode is considered successful
if the agent takes a \texttt{STOP} action within 1m of any instance of the target object category within $500$ timesteps. 



\vspace{-1em}
\paragraph{Overview}
At each timestep $t$, the robot receives an RGB-D observation. The depth image is used to update a partial map of the environment, identifying both navigable areas and obstacles. Frontiers are extracted as potential exploration targets. To guide exploration, a VFM model (GPT-4o) is queried using the current RGB image and the target object. The model estimates the likelihood of finding the target object in the direction of the current observation. This score is associated with the frontier. A long-horizon global planner LSP\cite{stein2018learning}, selects the most promising frontier that minimizes the expected cost of navigation. A local navigation module then computes a low-level action to move the agent toward the selected frontier. Once the target object is detected, the local navigation module guides the robot to its location and issues a \texttt{STOP} action (\cref{fig:overview}). 

\subsection{Mapping module}
We maintain an egocentric, partial map of the environment that is updated incrementally as the agent explores. At each timestep, the depth image is converted into a 3D point cloud, from which free space and obstacles are identified using a height threshold. Frontiers are extracted as boundaries between free and unknown space. For each frontier, the center point along its boundary is selected as the navigation target.

In addition to the geometric map, we also construct a two-channel value map, inspired by VLFM \cite{yokoyama2023vlfm}, consisting of a semantic value, $V$ and a confidence score map, $C$. The semantic value map stores the estimated likelihood of finding the target object from the VLM (\cref{sec:VLM}) in each explored cell.  Instead of averaging VLM scores across timesteps, we update the value map using a confidence score map that reflects the agent’s visual certainty. Confidence is highest (1) along the optical axis and decays to 0 at the edges of the field of view (FOV), ensuring that semantic scores are updated more reliably in well-observed regions. The map cells corresponding to the current FOV are updated using the  equations: $v_{i,j}^{new}=((c_{i,j}^{curr}v_{i,j}^{curr})+(c_{i,j}^{prev}v_{i,j}^{prev})/(c_{i,j}^{curr}+c_{i,j}^{prev}); c_{i,j}^{new}=((c_{i,j}^{curr})^2 + (c_{i,j}^{prev})^2)/(c_{i,j}^{curr}+c_{i,j}^{prev})$

\subsection{Assigning VLM-probabilities to frontiers}
\label{sec:VLM}
To guide exploration, we leverage GPT-4o to estimate the likelihood of encountering the target object if the agent moves in the direction of its current observation. At each timestep $t$, the RGB image $I_t$ is fed into GPT-4o along with the prompt:
``Output only a floating point value denoting the likelihood of finding a $\textlangle \text{target object}\textrangle$ if I move in this direction." Unlike prior approaches that rely on image captioning models (e.g., BLIP-2) \cite{yokoyama2023vlfm} or spatial co-occurrence heuristics based on object-to-object relationships \cite{zhou2023esc, yu2023l3mvn}, GPT-4o provides a richer understanding of scene semantics. Its integrated vision-language reasoning capabilities allow it to infer plausible object locations by incorporating knowledge about indoor layouts, typical room-object associations, and commonsense spatial relationships. This makes it particularly effective in situations where the target object is not directly visible—such as when it is located through a doorway, or further down a hallway—where purely visual similarity metrics may fail (\cref{fig:ours>blip2}). Some responses from GPT-4o are included in the Appendix \cref{fig:p_scores}.
\vspace{-1em}
\begin{figure}[th]
    \centering
    \includegraphics[width=\linewidth]{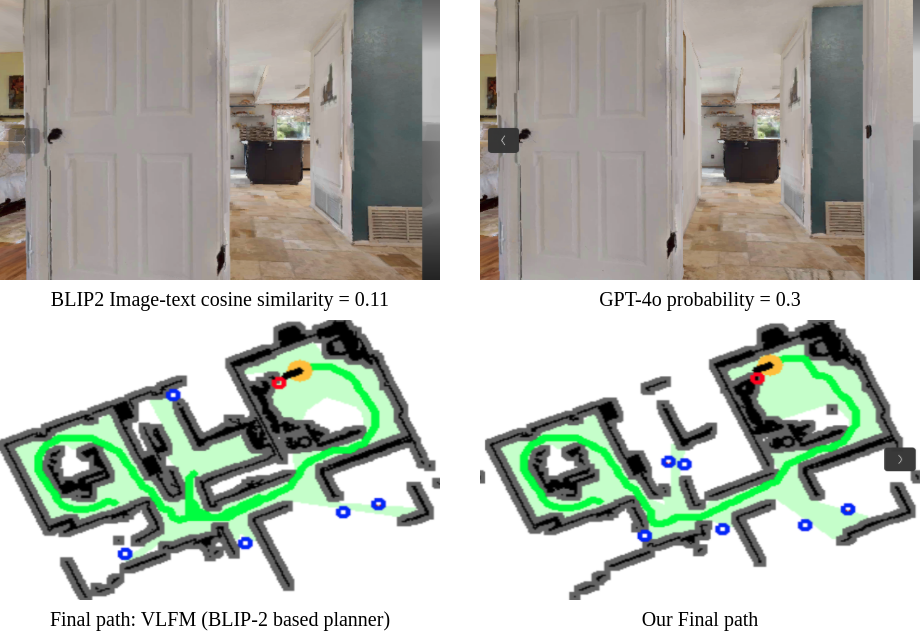}
    \caption{BLIP-2 assigns a lower score for the couch from the same location, leading the agent to explore the left room first, while our GPT-4o-based approach assigns a higher score and guides the agent directly through the hallway to find the couch.}
    \label{fig:ours>blip2}
\vspace{-1em}
    
\end{figure}

\subsection{Global planning with LSP}
Once the semantic scores for the frontiers are computed, we must select the most promising frontier. To do so, we adopt the LSP \cite{stein2018learning}, a model-based planning abstraction for efficient navigation. This approach has been shown to be effective for long-horizon reasoning in both exploration \cite{li2023learning} and point-goal navigation \cite{li2022comparison} tasks within simulated environments.

Unlike myopic methods that select the frontier with the highest immediate score, LSP evaluates the long-term utility of each frontier by estimating the expected cost of navigation as:

\vspace{-1em}
\begin{equation}
\label{eq:dp}
\begin{split}
Q(&\left\{m_t, q_t, \mathcal{A}_t \right\},a_t) = D(m_t, q_t, a_t) + P_SR_S +\\
& (1-P_S) (R_E + \max_{a\in \mathcal{A}_t \setminus \{a_t\}} Q(\left\{m_t, q_{a_t},  \mathcal{A}_t \! \setminus \! \{a_t\} \right\}, a)
\end{split}
\end{equation}

Here, $m_t$ is the current partial map, $q_t$ represents the agent's location, and $\mathcal{A}_t$ is the set of candidate frontiers. $Q(\left\{m_t, q_t, \mathcal{A}_t \right\},a_t)$ estimates the expected cost of successfully locating the target object by first navigating to frontier $a_t$. $D(m_t, q_t, a_t)$ is the distance from the agent to the frontier. $P_S$ is the probability of successfully finding the target object through frontier $a_t$, which we get from the VLM. $R_S$ and $R_E$ denote the success and exploration costs respectively.  Specifically, 
$R_S$ represents the expected distance to the object beyond a frontier if found, while 
$R_E$ captures the additional exploration and return cost if the object is not found. Since VFMs cannot reliably predict distances in unexplored regions, we assume $R_S=3m$ and $R_E=6m$ in our experiments, assuming unsuccessful exploration costs roughly twice as much as success.  Empirically, we observed that varying these constants has minimal impact on performance. We calculate the expected cost for all frontiers and select the one with the minimum expected cost.

\subsection{Local navigation and object detection}
The agent relies on a local navigation module to reach either the frontier or the target object location once it is visually detected by YOLOv7 \cite{wang2023yolov7}. Upon detection, we use Mobile-SAM to generate a segmentation mask of the object. The segmentation mask is then combined with the corresponding depth image to estimate the closest point on the object surface relative to the robot’s current position, which serves as the navigation goal. For reaching any target location (frontier or goal), we utilize the pretrained local point-goal navigation policy from \cite{yokoyama2023vlfm}, which was trained on the HM3D training split. This policy takes as input the current depth image and the relative position of the goal point, and produces discrete action commands to navigate towards it.

%% file: sec/5_experiments.tex
\section{Experiments}
\paragraph{Dataset}

We evaluate our approach using the Habitat \cite{savva2019habitat} simulator with the HM3D-val dataset \cite{ramakrishnan2021hm3d}, which comprises 2000 episodes across 20 diverse indoor scenes and includes navigation tasks for 6 object categories (bed, couch, chair, potted plant, toilet, tv).

\vspace{-1em}
\paragraph{Evaluation Metrics}

We report performance using two standard metrics: Success Rate (SR) and Success weighted by Path Length (SPL). SR measures the percentage of episodes in which the agent successfully navigates to an instance of the target object. SPL evaluates path efficiency by comparing the length of the agent’s trajectory to the shortest path from the starting location to the nearest instance of the target object.

\vspace{-1em}
\paragraph{Baselines}

We compare our approach against the following baselines:

\begin{itemize} \item \textbf{ZSON} \cite{majumdar2022zson}: Transfers a trained policy on image-goal navigation to semantic object-goals.

\item \textbf{ESC} \cite {zhou2023esc}: Uses LLM to reason about room-to-target object and discovered objects-to-target object relationship.

\item \textbf{VLFM} \cite {yokoyama2023vlfm}: It uses BLIP-2 \cite{li2023blip2} to compute cosine similarity between the current observation and the target object and uses the similarity score to prioritize frontiers.

\item \textbf{VoroNav} \cite {wu2024voronav}: Introduces a voronoi graph-based topological map, selecting subgoals using LLM-inferred probabilities with efficiency and exploration rewards.

\item \textbf{L3MVN} \cite {yu2023l3mvn}: Uses masked language model to assess the relevance between frontiers with nearby objects to target 

\item \textbf{TriHelper} \cite{zhang2024trihelper}: Builds on L3MVN by incorporating three manual intervention helpers—collision, exploration, and detection—to boost navigation performance.
\item \textbf{MFNP} \cite{zhang2024multi}: Supports exploration through staircases.

\end{itemize}

%% file: sec/6_result.tex
\section{Result}
\begin{table}[ht]
\centering
\setlength{\tabcolsep}{12pt}
{
\begin{tabular}{ccc}
\toprule
\multirow{2}{*} {\textbf{Approach}} & \multicolumn{2}{c}{\textbf{HM3D}} \\ \cline{2-3} 
                  & SPL $\uparrow$                 & SR$\uparrow$         \\ \midrule
ZSON \cite{majumdar2022zson}         & 12.6                 & 25.5       \\
ESC \cite{zhou2023esc}           & 22.3                 & 39.2       \\
VLFM \cite{yokoyama2023vlfm}      & {30.4}        & 52.5       \\ 
VoroNav \cite{wu2024voronav} & 26 & 42 \\ 
L3MVN \cite{yu2023l3mvn} & 23.1 & 50.4 \\ 
TriHelper \cite{zhang2024trihelper} & 25.3 & 56.5 \\ 
MFNP \cite{zhang2024multi} & 26.7 & \textbf{58.3} \\\midrule\midrule
SemNav-Greedy & {34.3} & 54.6 \\
SemNav & \textbf{35.9} & 54.9\\ \bottomrule
\end{tabular}%
}
\caption{Obj-nav results on HM3D validation set. SemNav-Greedy uses our frontier scoring mechanism with VLM and greedily selects the frontier with the highest probability score. SemNav uses LSP along with VLM-based scoring.}
\label{tab:results}
\end{table}

\begin{figure}[th]
    \centering
    \includegraphics[width=\linewidth]{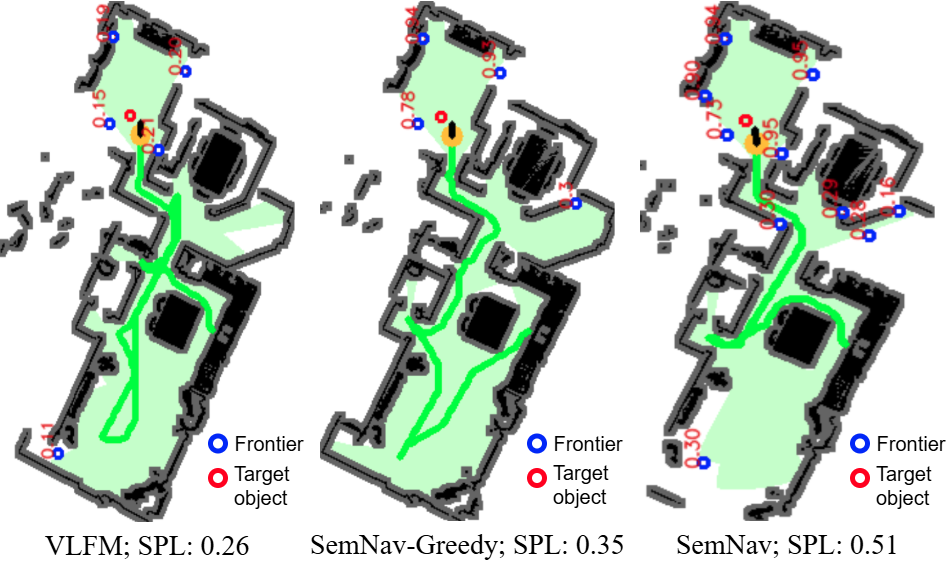}
    \caption{Final navigation paths for the target object bed. }
    \label{fig:ours>ourswLSP>blip2}
\vspace{-1em}
    
\end{figure}

\vspace{-1em}
Our method, SemNav, outperforms all baselines in terms of SPL, achieving the most efficient navigation paths (\cref{tab:results}). Compared to VLFM, which also leverages VLM to score frontiers, SemNav achieves $5.5\%$ and $2.4\%$ improvement in SPL and SR, respectively. These results show that our way of prompting the VLM for semantic reasoning, combined with long-horizon planning through LSP, leads to more effective exploration than relying solely on image-text cosine similarity as done in VLFM (\cref{fig:ours>blip2}). We observe a $1.6\%$ SPL drop with greedy frontier selection. (SemNav-Greedy vs SemNav in Table~\ref{tab:results}). \cref{fig:ours>ourswLSP>blip2} compares VLFM, SemNav, and SemNav-Greedy, showing that our approach achieves the highest SPL (detailed discussion in Appendix \cref{fig:vlfm_semNavGreedy_SemNav}). Specific examples illustrating SemNav’s behavior—such as efficiently exiting rooms when the target object is absent compared to VLFM is provided in the Appendix \cref{fig:door}. 

While our method significantly outperforms TriHelper and MFNP in SPL, it achieves slightly lower SR—by $1.6\%$ and $3.4\%$ respectively. These methods specifically address certain failure cases: TriHelper manually intervenes when the robot becomes stuck by relocating it to a more open area and double-checks object detections using a VLM, while MFNP further enables exploration through staircases. In contrast, our approach prioritizes improved semantic exploration through better frontier reasoning and does not currently handle stair navigation.

We also observed that around $25\%$ of the failed episodes are due to a simulator issue, where the agent successfully reaches an instance of the target object category but the simulator still registers the episode as a failure. This phenomenon has also been noted in previous works \cite{zhang2024trihelper, wasserman2024exploitation}. Representative examples illustrating this issue are provided in the Appendix (\cref{fig:simulator}).

%% file: sec/7_conclusion.tex
\section{Conclusion}
We proposed SemNav, a frontier-based zero-shot object-goal navigation framework that combines the semantic understanding of VLMs with model-based long-horizon planning. By leveraging VLM-driven frontier scoring and the LSP planner, our approach achieves state-of-the-art SPL performance on the HM3D dataset. Our results highlight the importance of general semantic reasoning and model-based planning for open-world embodied AI tasks.

%% file: sec/X_suppl.tex
\clearpage
\setcounter{page}{1}
\maketitlesupplementary

\section{Probability outputs from GPT-4o}
\label{sec:appendix}
\begin{figure}[h]
    \centering
    \captionsetup{type=figure, position=below}
    \includegraphics[width=0.95\textwidth]{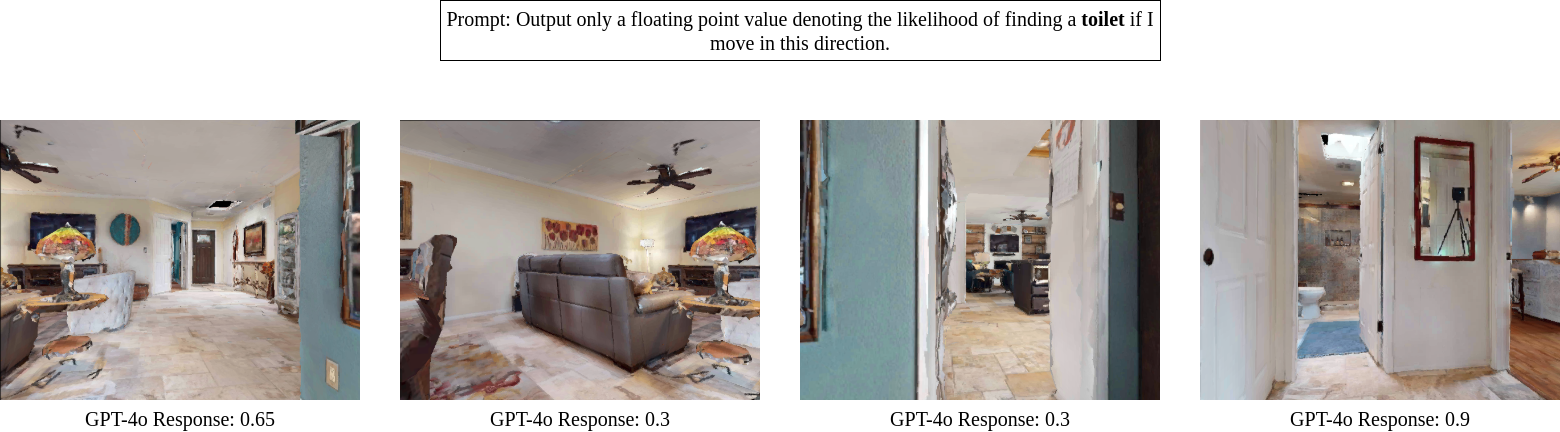}
    \parbox{\textwidth}{\caption{Some example images with their corresponding GPT-4o responses.}}
    \label{fig:p_scores}
    
\end{figure}

\section{Comparison between SemNav and VLFM}
\label{sec:door}
\begin{figure}[h]
    \centering
    \captionsetup{type=figure, position=below}
    \includegraphics[width=0.95\textwidth]{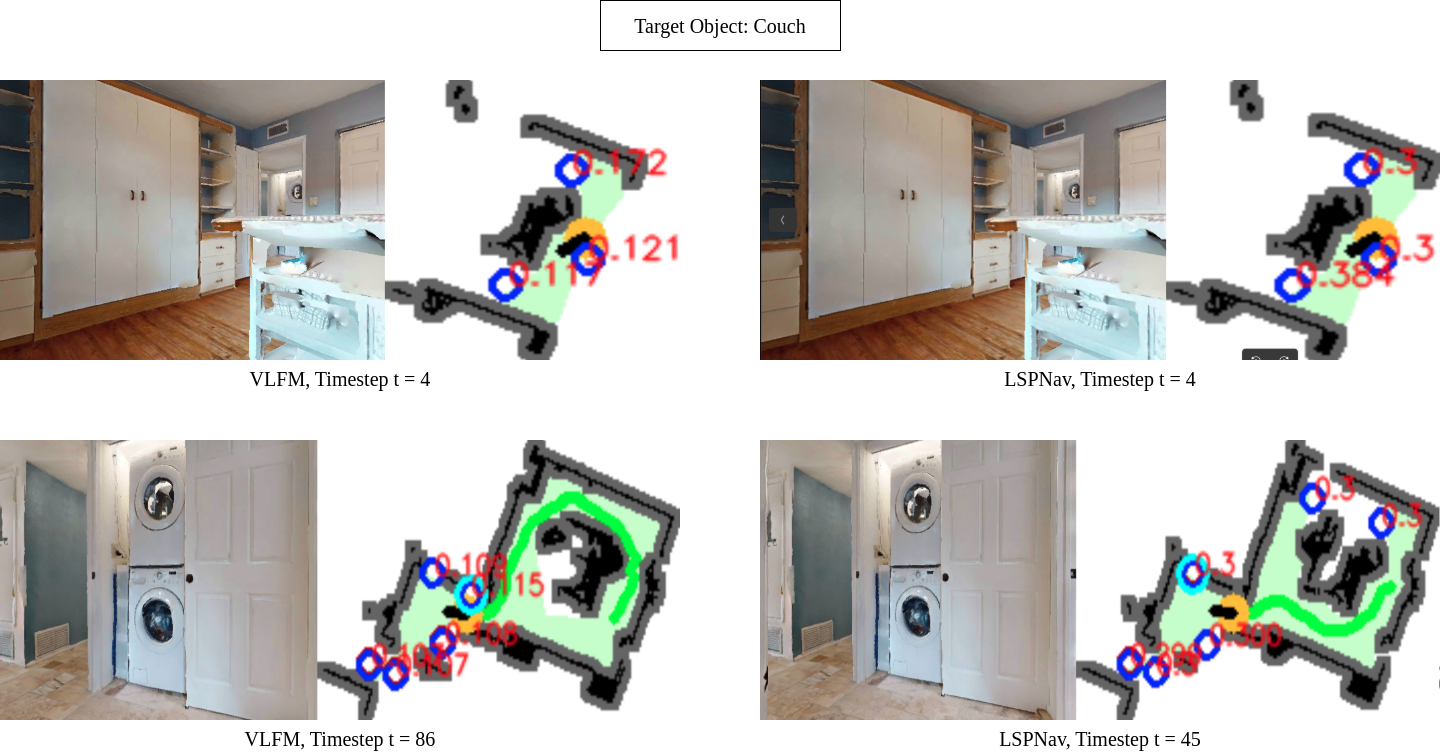}
    \parbox{\textwidth}{\caption{This is an example where we see that SemNav quickly gets out of the room to find a couch than VLFM. Top row: At timestep t=4, both are looking at a doorway which can lead to finding a couch. At that position SemNav had a higher probability score for the frontier closest to the doorway, which is evident from the map. This eventually leads SemNav agent to quickly get out of the room. Bottom row: SemNav gets out of the room at timestep $t=45$ whereas VLFM took timestep $t=86$ to get out of the room. The map shows a longer travelled path for VLFM.}}
    \label{fig:door}
    
\end{figure}
\clearpage

\section{Comparison among VLFM, SemNav-Greedy and SemNav}
\label{app:vlfm_semNavGreedy_SemNav}
\begin{figure}[h]
    \centering
    \captionsetup{type=figure, position=below}
    \includegraphics[width=0.95\textwidth]{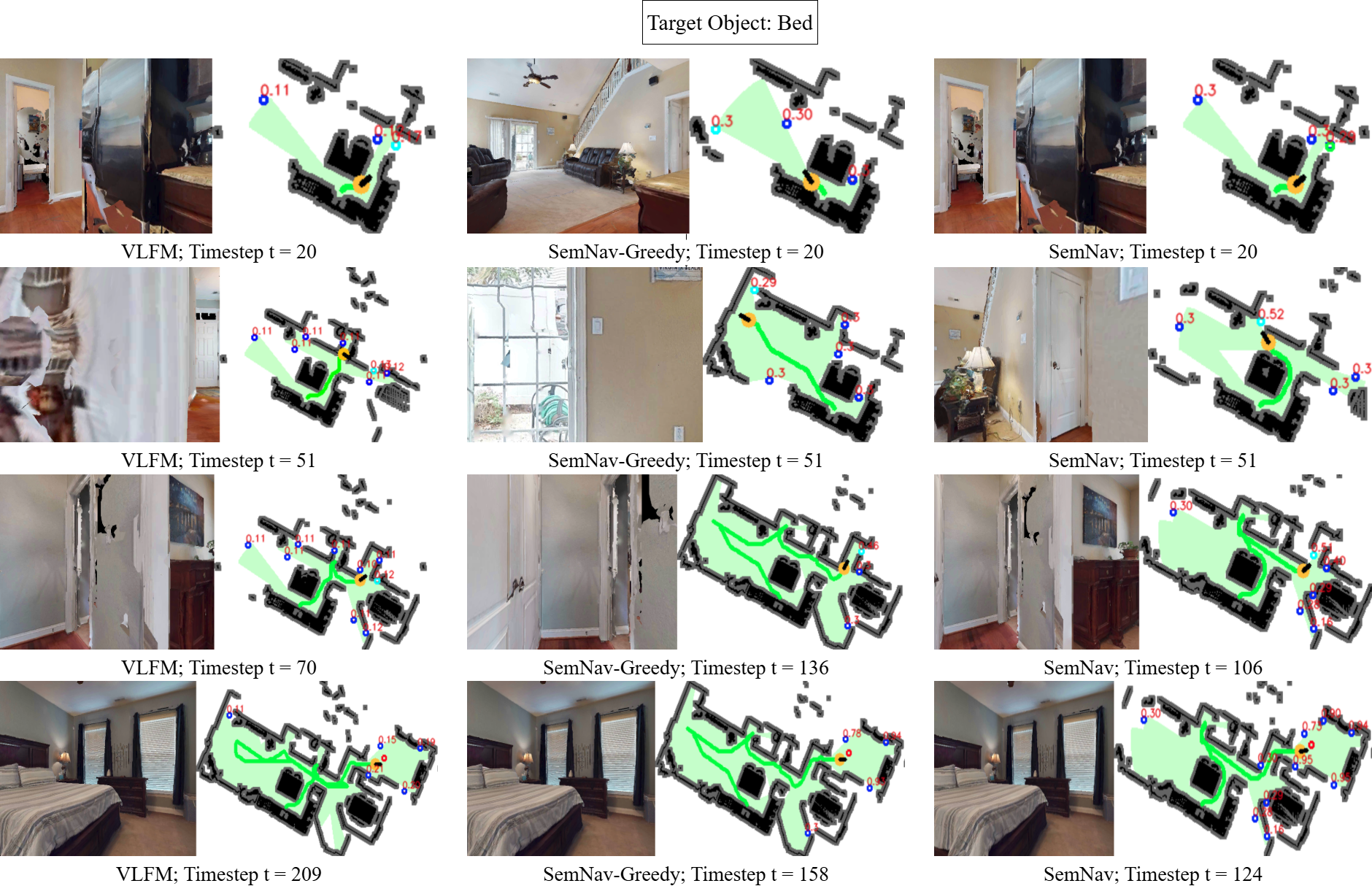}
    \parbox{\textwidth}{\caption{Row 1: At timestep $t=20$, SemNav (with LSP) selects a frontier with a lower score (0.29 in map) to minimize expected travel cost, choosing closer frontiers that guide it toward the target. In contrast, SemNav-Greedy selects the frontier with the highest score and moves in the opposite direction. Row 2: As a result, SemNav-Greedy is now farthest from the goal. Row 3: All agents observe the door leading to the goal. VLFM receives a poor score from BLIP-2 in that direction and continues exploring other frontiers first. SemNav and SemNav-Greedy, guided by higher GPT-4o scores, proceed directly through the door. Row 4: SemNav reaches the target object faster than all other approaches.}}
    \label{fig:vlfm_semNavGreedy_SemNav}
    
\end{figure}

\section{Simulator issue}
\label{sec:simulatorIssue}
\begin{figure}[h]
    \centering
    \captionsetup{type=figure, position=below}
    \includegraphics[width=0.95\textwidth]{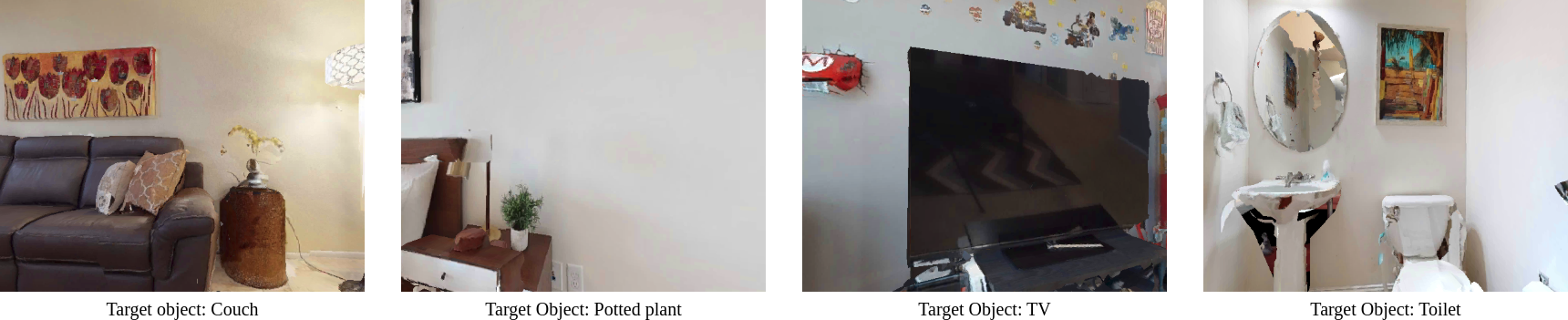}
    \parbox{\textwidth}{\caption{Here, we show a problem with the simulator where the agent successfully reached the target object, but it was still considered failed episodes.}}
    \label{fig:simulator}
    
\end{figure}